\documentclass[10pt,twocolumn,letterpaper]{article}

\usepackage{cvpr}              

\usepackage[pagebackref,breaklinks,colorlinks,allcolors=cvprblue]{hyperref}

\usepackage[utf8]{inputenc}
\DeclareUnicodeCharacter{2206}{$\Delta$}
\DeclareUnicodeCharacter{00E9}{\'e}  
\DeclareUnicodeCharacter{00FC}{\"u}  

\usepackage{multirow}
\usepackage{multicol}
\usepackage{algorithm}
\usepackage{algorithmic}
\usepackage{tcolorbox}
\usepackage{ulem}
\usepackage{soul}
\usepackage{marvosym}

\usepackage[accsupp]{axessibility}  

\definecolor{cvprblue}{rgb}{0.21,0.49,0.74}
\def\paperID{6450} 
\def\confName{CVPR}
\def\confYear{2025}

\title{\textit{Attend to Not Attended}: \\
Structure-then-Detail Token Merging for Post-training DiT Acceleration}

\author{Haipeng Fang$^{1,2}$,~Sheng Tang$^{1,2}$,~Juan Cao$^{1,2}$,~Enshuo Zhang$^{1,2}$,~Fan Tang\textsuperscript{\Letter,}$^{1,2}$, Tong-Yee Lee$^{3}$\\
$^{1}$ Institute of Computing Technology, Chinese Academy of Sciences \\
$^{2}$ University of Chinese Academy of Sciences\\
$^{3}$ National Cheng-Kung University\\
{\tt\small }
}

\begin{document}
\maketitle
\renewcommand{\thefootnote}{}
\footnotetext{\textsuperscript{\Letter}Corresponding author: Fan Tang.}
\begin{abstract}
Diffusion transformers have shown exceptional performance in visual generation but incur high computational costs. 
Token reduction techniques that compress models by sharing the denoising process among similar tokens have been introduced. 
However, existing approaches neglect the denoising priors of the diffusion models, leading to suboptimal acceleration and diminished image quality. 
This study proposes a novel concept: \textbf{attend to} prune feature redundancies in areas \textbf{not attended} by the diffusion process. 
We analyze the location and degree of feature redundancies based on the structure-then-detail denoising priors. 
Subsequently, we introduce \texttt{SDTM}, a structure-then-detail token merging approach that dynamically compresses feature redundancies. 
Specifically, we design dynamic visual token merging, compression ratio adjusting, and prompt reweighting for different stages. 
Served in a post-training way, the proposed method can be integrated seamlessly into any DiT architecture. 
Extensive experiments across various backbones, schedulers, and datasets showcase the superiority of our method, for example, it achieves 1.55$\times$ acceleration with negligible impact on image quality. Project page: \url{https://github.com/ICTMCG/SDTM}.

\end{abstract}    
\vspace{-3pt}
\section{Introduction}
\label{sec:introduction}
\begin{figure}[t] 
    \centering
	\includegraphics[width=1\linewidth]{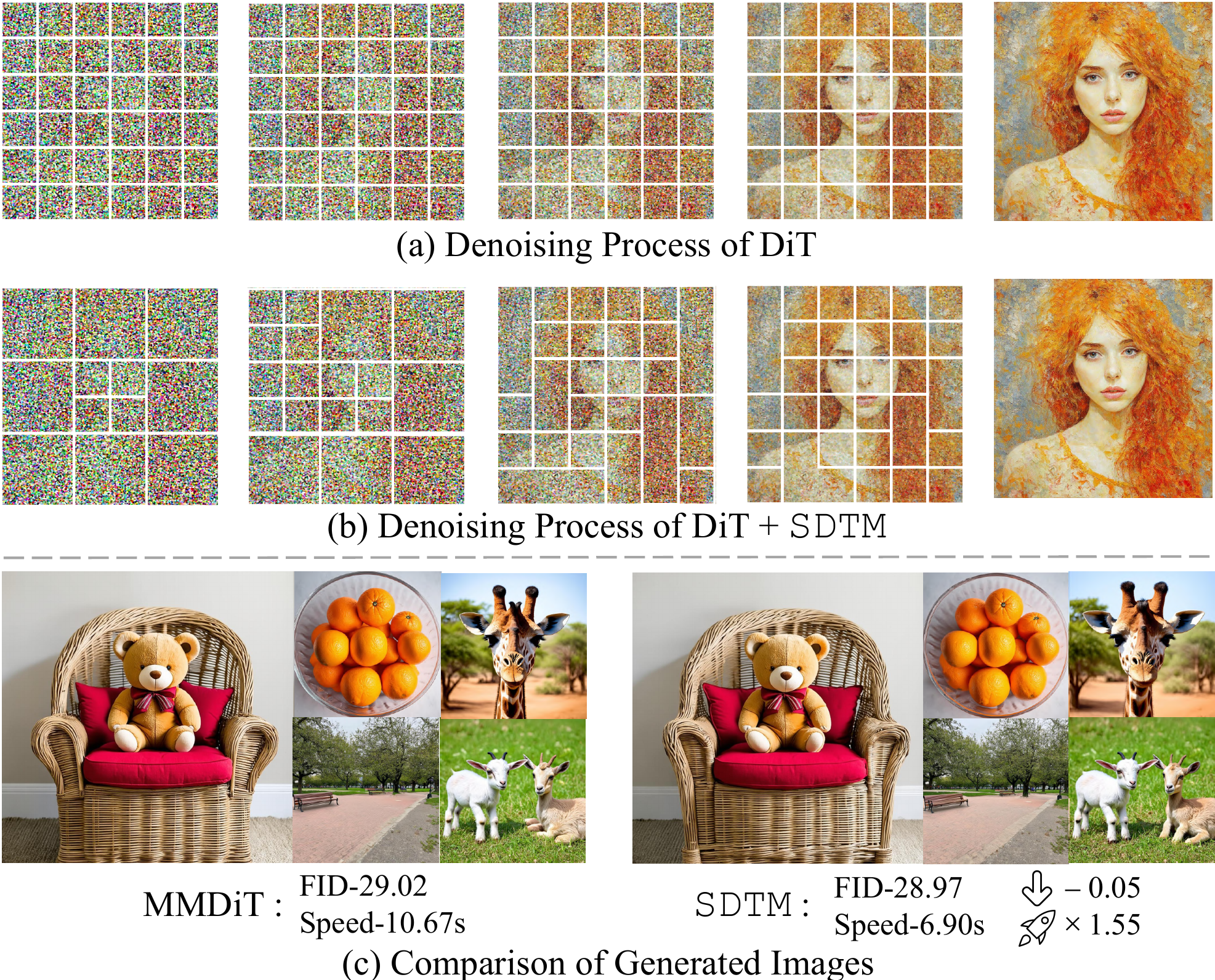}
        \vspace{-17 pt}
	\caption{Illustration. \textit{Upper}: Our \texttt{SDTM} represents a dynamic multi-resolution generation process by reducing feature redundancies in areas not unattended by the denoising process. \textit{Lower}: Compared to the baseline method, our approach achieves 1.55$\times$ acceleration with negligible impact on generation quality.} 
	\label{fig:motivation}
    \vspace{-10 pt}
\end{figure}
Diffusion transformers (DiTs)~\cite{DiT} are flourishing in image~\cite{pixart-a, mmdit, FLUX} and video~\cite{latte, opensora, open_sora_plan} generation, and have been adopted as the fundamental model of Sora~\cite{sora}. 
However, heavy computational redundancies slow down inference and drive the need for acceleration techniques.

Several sampler-based methods aim to optimize denoising steps through sampler optimization~\cite{DDIM, DPMSolver} or distillation~\cite{salimans2022progressive, meng2023distillation}, and model-based studies focus on pruning~\cite{fang2024structural,castells2024ld}, quantizing~\cite{Q-diffusion, ptqd}, or caching~\cite{ma2024deepcache, wimbauer2024cache} architectural redundancy. Due to the inflexibility of these methods for diverse data and evolving requirements, various studies introduced token reduction~\cite{rao2021dynamicvit, evit, tokenmerge, norouzi2024algm} to reduce feature redundancy. For instance, ToMeSD~\cite{ToMeSD} and AT-EDM~\cite{ATEDM} compress similar or unimportant tokens, where DyDiT~\cite{dydit} and TokenCache~\cite{tokencache} train a dynamic module to prune or cache unnecessary computation. However, these feature-based approaches often require additional fine-tuning or overlook the denoising priors, resulting in suboptimal acceleration, diminished image quality, and limited applicability.

In this paper, we propose a novel view: \textbf{\textit{attend to}} prune feature redundancies in areas \textbf{\textit{not attended}} by the diffusion process. Initially, we observe the evolution of low and high frequencies within DiT and confirm that diffusion transformers still adhere to the structure-then-detail denoising priors: they allocate less attention to less-structure tokens in the early steps and to weak-detail tokens in the later steps. Subsequently, we hypothesize that these unattended tokens may be redundant, and we validate this hypothesis by tracking the location of feature redundancies. Furthermore, we find that the degree of feature redundancy in the early diffusion process is significantly greater than in the later stages.

Based on the above priors, we introduce \texttt{SDTM}, a novel structural-then-detail token merging approach that \textbf{\textit{dynamically}} reduces token redundancies stepwise without requiring additional fine-tuning. For visual token merging, we develop a similarity-prioritized structural merging method, an inattentive-prioritized detail merging method, supplemented by time-wise ratio adjusting to alter the compression degree at different stages. From the perspective of prompt guidance, we design a time-wise prompt reweighting to optimize the guidance direction at different steps. In summary, the contributions of this paper are as follows:

\begin{itemize}
    \item We analyze the location and degree of feature redundancies and introduce \texttt{SDTM}, a dynamic token compression approach that employs a structure-then-detail token merging strategy. It is finetuning-free and can be seamlessly integrated into any text-to-image DiT architecture.
    \item We design similarity-prioritized structural merging and inattentive-prioritized detail merging methods for different stages, supplemented by time-wise ratio adjusting and prompt reweighting to compress feature redundancies.
    \item Quantitative and qualitative experiments across multiple backbones, schedulers, and datasets demonstrate our method's superiority. For example, \texttt{SDTM} achieves 1.55$\times$ acceleration with negligible impact on generation quality.
\end{itemize}
\section{Related Work}
\label{sec:relatedworks}
\subsection{Diffusion Transformers}

Diffusion models (DMs)~\cite{DDPM, ADM, Score-based} transform noise into complex data distributions via reversible Markov processes. Early U-Net based DMs achieved remarkable results in image~\cite{LDM, podell2023sdxl, zhang2023adding} and video generation~\cite{singer2022make, chen2023videocrafter1, zhou2022magicvideo}. Currently, diffusion transformers (DiTs)~\cite{DiT} are gaining prominence due to their scalability. Building on DiTs, image generation models such as PixArt-$\alpha$~\cite{pixart-a}, MMDiT~\cite{mmdit}, and FLUX~\cite{FLUX} have been introduced. Furthermore, DiTs have been recognized as a fundamental component in the Sora~\cite{sora}, leading to the development of video generation models~\cite{latte, opensora, open_sora_plan}. However, significant computational redundancies of diffusion sampling process limit its widespread application.

\subsection{Efficient Diffusion Models}
Numerous efforts have been proposed to compress the redundancies in the denoising process via samplers, architectures, and feature computation, it can be categorized as:

\textit{Sampler-based:} Various approaches focus on optimizing samplers; for instance, DDIM~\cite{DDIM} offers a non-Markovian variant of the diffusion process, DPMSolver~\cite{DPMSolver} applies a numerical solver to the differential equations and Rectified Flow~\cite{rectifiedflow} optimizes distribution transport in ODE models. Additionally, several progressive distillation techniques~\cite{salimans2022progressive, meng2023distillation, kim2024distilling} attempt to distill the sampler into fewer steps. 

\textit{Model-based:} Some studies focus on reducing architecture redundancy: methods like Diff-pruning~\cite{fang2024structural,kim2023bk,castells2024ld,li2024snapfusion} advocate pruning unimportant weights; quantization methods represented by Q-diffusion~\cite{Q-diffusion, ptqd, zhao2024vidit, so2024temporal} aim to quantize redundant modules. Furthermore, some caching mechanisms~\cite{ma2024deepcache,wimbauer2024cache,li2023faster,chen2024delta} enhance efficiency by caching and reusing module outputs across adjacent denoising steps.

\textit{Feature-based:} Given the dependency of sampler-based and model-based methods on fixed strategies, they struggle with varying data patterns and compression demands. Therefore, some approaches explore feature compression techniques. For instance, within the U-Net architecture, ToMeSD~\cite{ToMeSD} and AT-EDM~\cite{ATEDM} advocate for compressing similar or unimportant tokens, whereas in DiT architectures, DyDiT~\cite{dydit} and TokenCache~\cite{tokencache} train a dynamic token module to prune or cache unnecessary computation.

In contrast to other feature-based methods that utilize relatively \textit{static} reduction strategies or require fine-tuning, our \texttt{SDTM} is specifically designed to perform \textit{dynamic} post-training token reduction. This approach is based on thoroughly analyzing the location and degree of feature redundancies at different stages of diffusion process. Therefore, \texttt{SDTM} is more efficient compared to the above frameworks.

\section{Preliminary and Motivation}
\label{sec:pre}
\begin{figure*}[t] 
    \centering
	\includegraphics[width=0.9\linewidth]{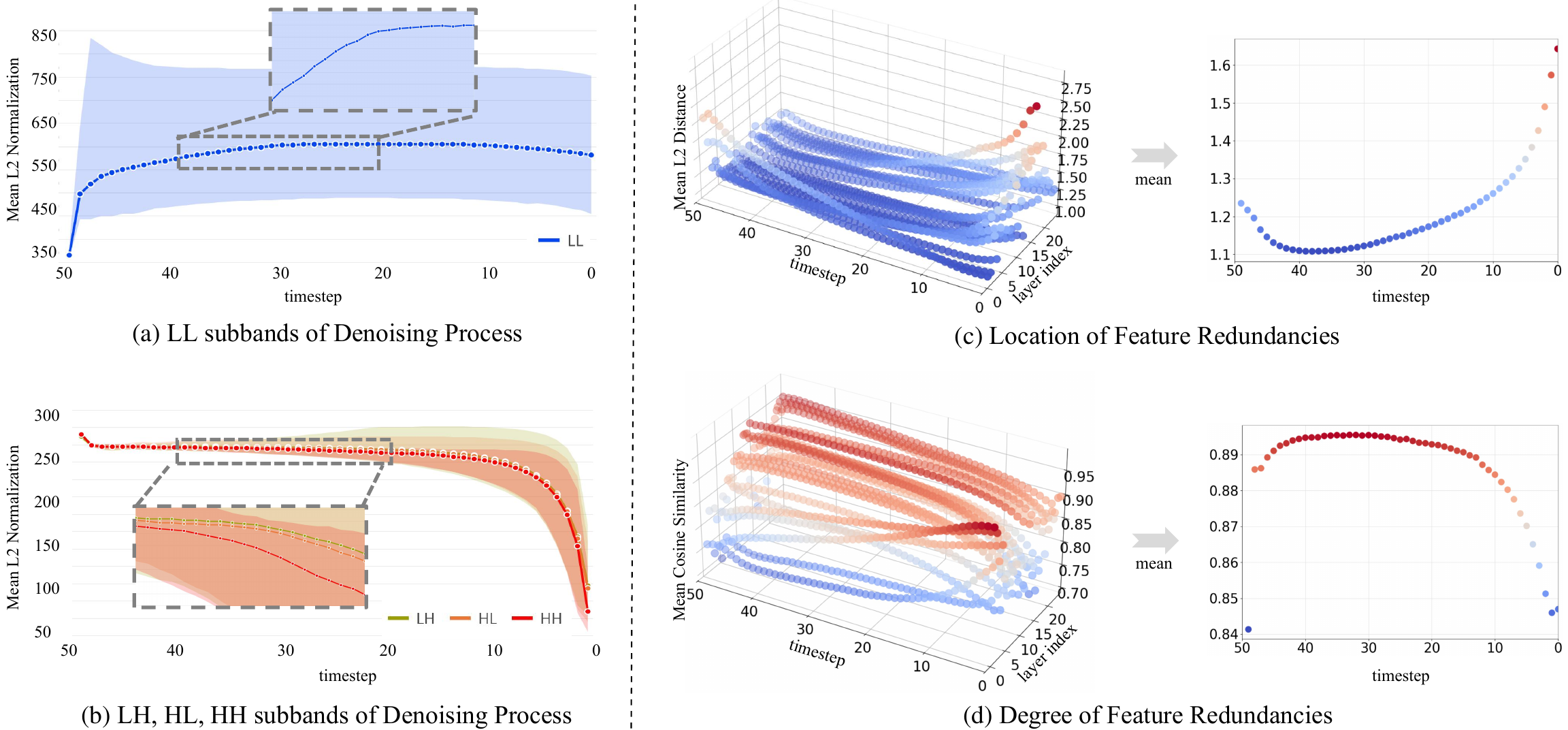}

	\caption{Preliminaries. \textit{Left:} Evolution of Denoising Process: The L2 norm of (a) LL, (b) LH, HL, and HH subbands of estimated noise during the DiT denoising process. \textit{Right:} Evolution of Feature Redundancies: (c) Location and (d) Degree evolution of token redundancies across DiT different steps and layers. Experiments were conducted on MMDiT~\cite{mmdit} with 50-step RF schedule~\cite{rectifiedflow} based on 10k samples.} 
    \vspace{-10 pt}
	\label{fig:pre}
\end{figure*}

Our approach is driven by a straightforward idea: \textit{attend to} prune feature redundancies in areas \textit{not attended} by the diffusion process. We assume that tokens unattended by the diffusion process might be redundant, and raise two following questions. Experiments were conducted on MMDiT~\cite{mmdit} with 50-step RF schedule~\cite{rectifiedflow} based on 10k samples.

\noindent \textit{\textbf{Which ones are not attended by denoising process?}} 
A well-known prior of the denoising process is that it includes a structure stage for planning visual semantics, followed by a detail stage that enhances the visual fidelity~\cite{DPMSolver,yang2023diffusion,qian2024boosting}. From this perspective, we aim to verify whether this prior still holds in DiT. Specifically, we analyzed the evolution of low-frequency and high-frequency components of the estimated noise using DWT. As illustrated in Figs.~\ref{fig:pre} (a) and (b), DiT's architecture and scheduler have stabilized the frequency evolution, which fluctuated significantly in DDPM noted at~\cite{qian2024boosting}. The low-frequency bands show sharp changes in the first 40\% steps, with the high-frequency bands undergoing alterations in the subsequent 60\%. This pattern confirms that DiT maintains the structure-then-detail denoising process, highlighting that \ul{\textit{less-structure tokens in early steps and weak-detail tokens in later steps are the unattended ones}} which may be the redundant feature. 

\noindent \textit{\textbf{How to find these unattended ones?}} Firstly, we locate these unattended tokens and assess their redundancy. Specifically, we calculated the mean L2 distance and cosine similarity between the most similar pairs. Higher similarity indicates that they can substitute for each other which represents more redundancies~\cite{ToMeSD}, while the distance pinpoints their location. As illustrated in Fig.~\ref{fig:pre} (c), the distance decreases slightly and then increases progressively. Since higher similarity at lower distances implies less local structure, this identifies the initial locations of unattended tokens. Meanwhile, decreasing similarity indicates increasing detail diversity, with unattended tokens distributed among less-detail global areas. Therefore, \ul{\textit{the unattended tokens, which represent feature redundancies, initially cluster locally and gradually spread globally}}. Secondly, we analyzed the degree evolution of cosine similarity in Fig.~\ref{fig:pre} (d). Besides the initial low similarity among tokens immediately after the noise initialization, which quickly escalates after one step, the token similarity progressively decreases as timesteps, implying that token redundancy diminishes over time. Therefore, \ul{\textit{there are more unattended, redundant tokens in the early steps and fewer in the later steps.}}
\begin{figure*}[!t] 
    \centering
	\includegraphics[width=1\linewidth]{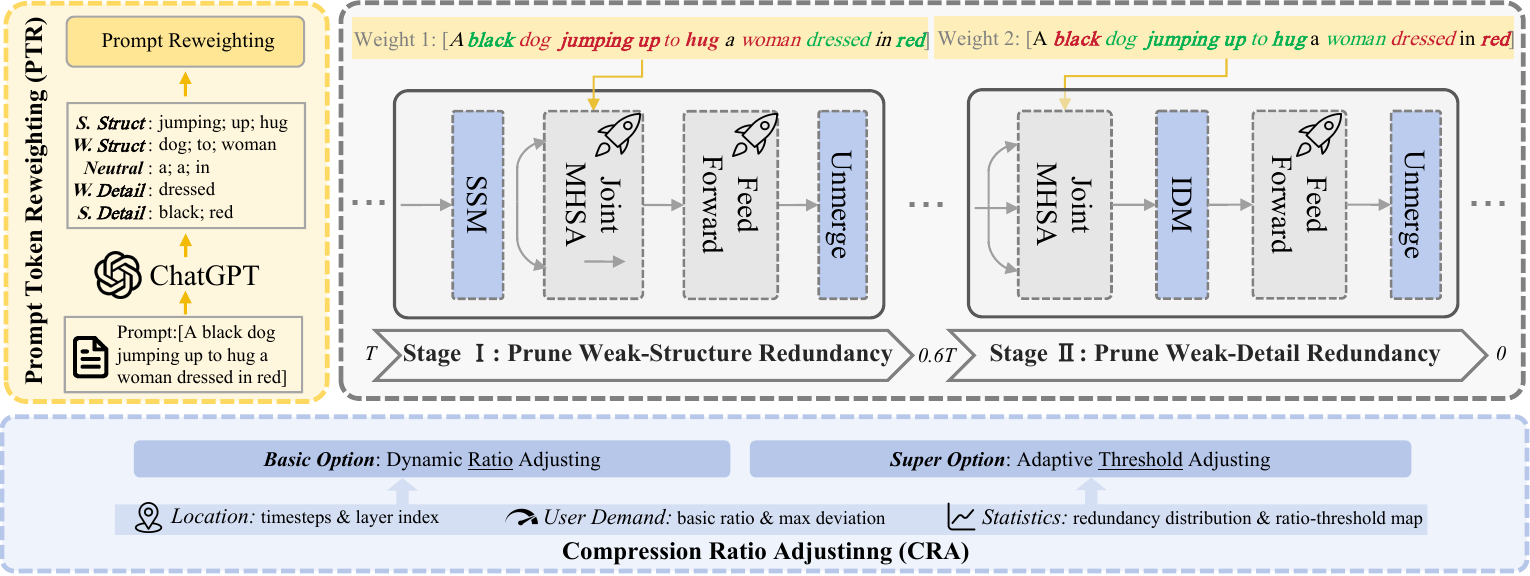}

	\caption{Overview. \textit{Grey:} Our \texttt{SDTM} compresses weak-structure redundancies in the early stage and weak-detail redundancies in the later stage. \textit{Blue:} Compression ratio adjusting (CRA) dynamically adjusts the ratio or threshold to control the pruning degree. \textit{Yellow:} Prompt token reweighting (PTR) categorizes each prompt token into structure or detail groups, optimizing the denoising direction by reweighting attention map. Here, increase and decrease in red and green, while bold implies intensity.} 
    \vspace{-6 pt}
	\label{fig:overview}
\end{figure*}
\begin{figure}[!t] 
    \centering
	\includegraphics[width=1\linewidth]{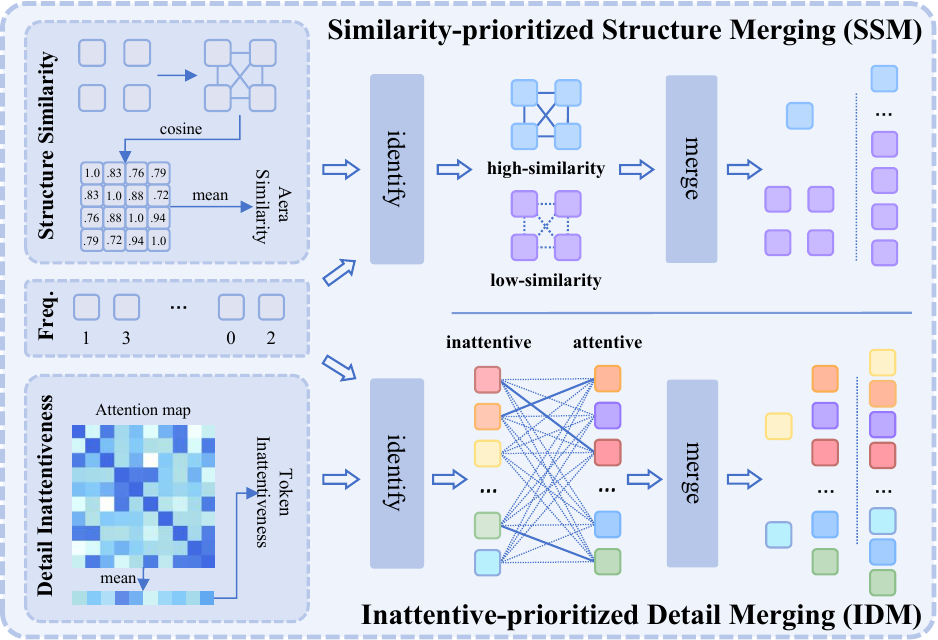}

	\caption{Visual Token Merging. By measuring structure similarity, unmerged frequency, and detail inattentiveness, SSM and IDM target different types of feature redundancies for reduction.} 
    \vspace{-6 pt}
	\label{fig:merge}
\end{figure}
\section{Methodology}

As shown in Fig.~\ref{fig:overview}, we introduce the \texttt{SDTM} framework, a structure-then-detail token merging approach that facilitates dynamic token compression to reduce feature redundancy. Based on the analysis in Sec.~\ref{sec:pre}, we perform DiT acceleration across two stages. At different stages, we adopt various visual token merging (shown in Fig.~\ref{fig:merge}) and prompt token reweighting methods, detailed in Sec.~\ref{sec:visual} and Sec.~\ref{sec:prompt}.

\subsection{Visual Token Merging}
\label{sec:visual}
To address feature redundancies vary in form and location, we design a similarity-prioritized structure merging (SSM) to address local redundancies in structure stage and an inattentive-prioritized detail merging (IDM) to reduce global redundancies in detail stage. Furthermore, considering the higher redundancies in early steps and lower in later steps, we develop compression ratio adjusting (CRA) that dynamically optimizes ratios or thresholds across timesteps. 

\subsubsection{Similarity-prioritized Structure Merging} 
Based on the insights derived from Sec.~\ref{sec:pre}, we reduce the weak-structure feature redundancy in local areas during the early stage. We developed similarity-prioritized structure merging and integrated them before the MHSA and MLP blocks. The SSM involves two processes: identifying high-redundancy tokens and merging them dynamically.

\noindent{\textbf{Identifying.}} In the initial stage, tokens with less structure are identified within local areas. The feature embedding is represented by \(\mathcal{X} \in \mathbb{R}^{N \times d}\), where $N=H\times W$ and \(H, W\) denote the height and width. Inspired by ALGM~\cite{norouzi2024algm}, we reshape \(\mathcal{X}\) into a grid \(\mathcal{X} \in \mathbb{R}^{m \times \frac{H}m \times m \times \frac{W}{m} \times d}\), with \(m \times m\) indicates the window size. By grouping the tokens of each window, we define \(\mathcal{X} = \{w_1, \ldots w_k, \ldots \}\). Subsequently, we compute the average cosine similarity within each window \(w\) as the similarity priority score $\mathcal{P}^{sim}_{w}$:
\begin{equation}
\mathcal{P}^{sim}_w = \frac{ \sum^{i \neq j }_{x_i, x_j \in w} \cos(x_i, x_j)}{m^2 \cdot (m^2 - 1)}.
\end{equation}
Additionally, we observed that error accumulation tends to escalate when a token is continuously merged. To mitigate this, we prioritize tokens unmerged recently over those involved in continuous merging events. We monitor the times since each token’s last merging operation, represented as $\mathcal{T}_x$, and calculate $\mathcal{P}^{fre}_x = \frac{\mathcal{T}_x}{\mu({\mathcal{T}})}$ for each token. We calculate frequency priority score $\mathcal{P}^{fre}_w$ for window \(w\):
\begin{equation}
    \mathcal{P}^{fre}_w = \frac{ \sum_{x_k\in w} \mathcal{P}^{fre}_{x_k}}{m^2}.
\end{equation}
Finally, we compute the total priority score \( \mathcal{P}_w = \mathcal{P}^{sim}_w + \alpha_s \mathcal{P}^{fre}_w \), where \( \alpha_s \) is a scaling factor. We sort the windows in descending order to prioritize the redundancies.

\noindent{\textbf{Merging.}} Our merging strategy can be controlled using either a ratio \( \rho \) or a threshold \( \theta \). We select either the top \( \rho \) redundant windows or windows where \( \mathcal{P}_w > \theta \) for merging. Both \( \rho \) and \( \theta \) are dynamically adjusted throughout the timesteps, as detailed in Sec.~\ref{sec:adjust}. Subsequently, we average the tokens within each selected window and retain the tokens from unselected windows to construct a new feature embedding \( \mathcal{X}' \in \mathbb{R}^{N' \times d} \). In this process, \( N' \) is reduced compared to \( N \), thereby decreasing the computational cost on the subsequent MHSA and MLP blocks.

\subsubsection{Inattentive-prioritized Detail Merging}
Following the analyses in Sec.~\ref{sec:pre}, we reduce weak-detail feature redundancy in later stages of the denoising process. We developed inattentive-prioritized detail merging and incorporated them before the MLP blocks. We have ceased accelerating the MHSA module, as its role in facilitating global information interactions is crucial, and its computational demand is significantly lower than that of MLP. Similarly, the IDM includes identifying and merging processes.

\noindent{\textbf{Identifying.}} We assume that tokens with minimal impact on others are information-sparse and thus weak-detail. Utilizing the attention map, which effectively quantifies the relationships between tokens, we identify the inattentive tokens. For attention map $\mathcal{A} \in \mathbb{R}^{N \times N}$, $\mathcal{A}(x_i,x_j)$ quantifies the influence of the $j$-th token on the $i$-th token, a notion widely recognized in~\cite{rao2021dynamicvit, liang2022not}. We then compute the inattentive priority score $\mathcal{P}^{ina}_{x}$ for each token as follows:
\begin{equation}
\mathcal{P}^{ina}_x = 1 - \frac{ \sum_{k \in 1...N } \mathcal{A}(x_k, x)}{N}.
\end{equation}
Subsequently, we calculate \( \mathcal{P}^{fre}_x \) for each token and its overall priority score \( \mathcal{P}_x = \mathcal{P}^{ina}_x + \alpha_d \mathcal{P}^{fre}_x \), where $\alpha_d$ is a scaling factor. We then sort the tokens in descending order to effectively prioritize the feature redundancies.

\noindent{\textbf{Merging.}} Based on the priority score, we categorize tokens into inattentive and attentive groups, as shown in Fig.~\ref{fig:merge}. We calculate the cosine similarity between these groups and identify the maximum cosine similarity \( \mathcal{S}_i \) for each token in inattentive group. Depending on the merging ratio \( \rho \) or threshold \( \theta \), we select either the top \( \rho \) redundant tokens or those which \( \mathcal{S}_i > \theta \) for merging. It is essential that \( \mathcal{P} \) determines the merging priority, whereas the merging process relies on \( \mathcal{S} \). Considering \( x_i \) and \( x_a \) as an example, given the differing information content between inattentive and attentive groups, the merged \( x_a' \) can be formulated as:
\begin{equation}
x'_a = softmax([1-\mathcal{P}^{ina}_{x_i}, 1-\mathcal{P}^{ina}_{x_a}])\cdot [x_i,x_a].
\end{equation}

\subsubsection{Compression Ratio Adjusting}
\label{sec:adjust}
The variability in the degree of feature redundancies across timesteps is evident, as shown in Fig.~\ref{fig:pre}. With increasing timesteps, there is a general decline in the degree of feature redundancies. In addition, specific steps (the first) and specific layers (the first four) deviate from this trend. In practical scenarios, users often have specific compression requirements, which typically include a basic ratio $\rho$ and a maximum deviation $d$. To meet these demands, we propose two strategies: dynamically adjusting the compression ratio and adaptively adjusting the compression threshold.

\noindent \textbf{Dynamic Ratio.} This method offers a straightforward approach for dynamically adjusting the merging ratio. It modifies the ratio to follow a cosine decay of $[0, \frac{\pi}{2}]$ from $\rho + d$ to $\rho - d$ across various steps,  in alignment with the cosine-shaped curve of feature redundancies illustrated in Fig.~\ref{fig:pre}. For specific steps and layers, the ratio is set directly to the minimum $\rho - d$. We adopt this strategy in our baseline \texttt{SDTM} model because it provides a close approximation of the optimal ratio without the need for complex computations. 

\noindent \textbf{Adaptive Threshold.} Due to the varying complexity of generating different images, employing a constant merging ratio may result in suboptimal acceleration for simpler images and compromised quality for more complex ones. To address this issue, we design an adaptive method to automatically adjust the threshold. We initially sampled a small image batch to assess the similarity across steps and layers to create a distribution $S$. Then, we constructed a ratio-threshold mapping table $M$ by documenting the average threshold at 1\% intervals of the merging ratio. The comprehensive process is outlined in Algorithm~\ref{alg:alg}. This approach has been integrated into our enhanced \texttt{SDTM*}, better suited for industrial-scale inference scenarios where neither suboptimal acceleration nor quality degradation is acceptable.

\begin{algorithm}
\caption{Adaptive Threshold Adjusting}
\label{alg:alg}
\begin{algorithmic}[1]
\REQUIRE similarity distribution $S$, basic ratio $r$, max deviation $d$, ratio-threshold map $M$

\textbf{Input:} timestep \& layer sequence $\{(1,1),...,(t,l)\}$

\textbf{Output:} adaptive threshold $\theta = \{\theta_{(1,1)},..., \theta_{(t,l)}\}$

\STATE Scale similarity distribution \( S \) to range \([-1, 1]\).

\FOR{each $t$}
    \FOR{each $l$}
        \STATE Compute $S_{t, l}$ for current $t$ and $l$
        \STATE $\rho_{(t,l)} \leftarrow r + d \cdot S_{(t, l)}$
        \STATE $\theta_{(t,l)} \leftarrow M(t,l,\rho_{(t,l)})$ 
    \ENDFOR
\ENDFOR
\end{algorithmic}
\end{algorithm}

\subsubsection{Token Unmerging}
Image generation is a feature-intensive task that necessitates the complete feature map. Using the example of \( x_i \) and \( x_a \), we need to use \( x_a' \) to obtain new \( x_i'' \) and \( x_a'' \). In related work such as ToMeSD~\cite{ToMeSD} and AT-EDM~\cite{ATEDM}, the strategy of similarity-based token reuse has been utilized, directly substituting \( x_a' \) for \( x_i'' \) and \( x_a'' \). However, this substitution can introduce replacement errors, particularly when employing a higher pruning ratio. Therefore, we initially employ similarity-based token reuse to restore \( x''(t) \) at the current timestep, followed by a weighted combination with the previous timestep's \( x''(t-1) \). In this method, \( x''(t-1) \) preserves the independence of individual token features, while \( x''(t) \) integrates the new timestep's denoising features.

\begin{table*}[t]
\centering
\resizebox{1\linewidth}{!}{
    \setlength{\tabcolsep}{11pt}{
    \renewcommand{\arraystretch}{1}
    
    \begin{tabular}{l | c | c c c  | c c | c }
        \toprule
        \multirow{2}*{\bf Method}  & \multirow{2}*{\bf Finetune}  & \multirow{2}*{\bf MACs(T) $\downarrow$} & \multirow{2}*{\bf Latency(s) $\downarrow$} & \multirow{2}*{\bf Speed $\uparrow$}  & \multicolumn{2}{c|}{\bf COCO2017} & \bf PartiPrompts \\
            & & & & & \bf FID $\downarrow$ & \bf CLIP $\uparrow$ & \bf CLIP $\uparrow$  \\
        \midrule 
      \textbf{SD3 Medium}  \cite{mmdit}      &  & 6.01  & 10.67   & {1.00$\times$} & 29.02          & 0.3267 & 0.3279\\
      \midrule
ToMeSD - a  & & 4.27 &   8.08 &  {1.32$\times$}  & 45.28 & 0.3102 &0.3108 \\
AT-EDM - a   & & 4.23 &   8.14 & {1.31$\times$} & 34.72 & 0.3195& 0.3219 \\
Ours-\texttt{SDTM} - a & & 4.20 & 8.20 & {1.30$\times$} & 28.73 & 0.3235 & 0.3248 \\
Ours-\texttt{SDTM}* - a& & \textbf{4.13} & \textbf{8.02} & \textbf{{1.33$\times$}} & \textbf{28.57}  & \textbf{0.3249} & \textbf{0.3261} \\
      \midrule
ToMeSD - b &&3.81  & 7.11 & {1.50$\times$} & 75.44 & 0.2780 & 0.2816\\
AT-EDM - b &&3.78 & 7.02 & {1.52$\times$}& 43.92 & 0.3089 & 0.3125\\
TokenCache - b   &\checkmark &3.72  &6.97 & {{1.53$\times$}}  &28.83  &0.3208 & 0.3226 \\
DyDiT - b  &\checkmark &3.74 &\textbf{6.87} & \textbf{{1.55$\times$}}   &\textbf{28.47}  &0.3213 & 0.3227\\
Ours-\texttt{SDTM} - b &&3.66  & 7.07 & {1.51$\times$}&29.60  &0.3224 &0.3231 \\
Ours-\texttt{SDTM}* - b &&\textbf{3.62} & 6.90 &\textbf{{1.55$\times$}}                       &28.97   &\textbf{0.3237} & \textbf{0.3252}\\

      \bottomrule
      
      \end{tabular}}
}
\vspace{-5 pt}
\caption{\textbf{Quantitative comparison} on MS-COCO2017 and PartiPrompts with $\text{Stable Diffusion 3 medium}$ and 50 steps rectified flow by default. For configurations a and b, we adjust the compression ratios of various methods to reach approximate speeds of 1.3$\times$ and 1.5$\times$.}
\label{table:sd3_comparison}
\vspace{-5 pt}
\end{table*}
\begin{figure*}[!t] 
    \centering
	\includegraphics[width=0.96\linewidth]{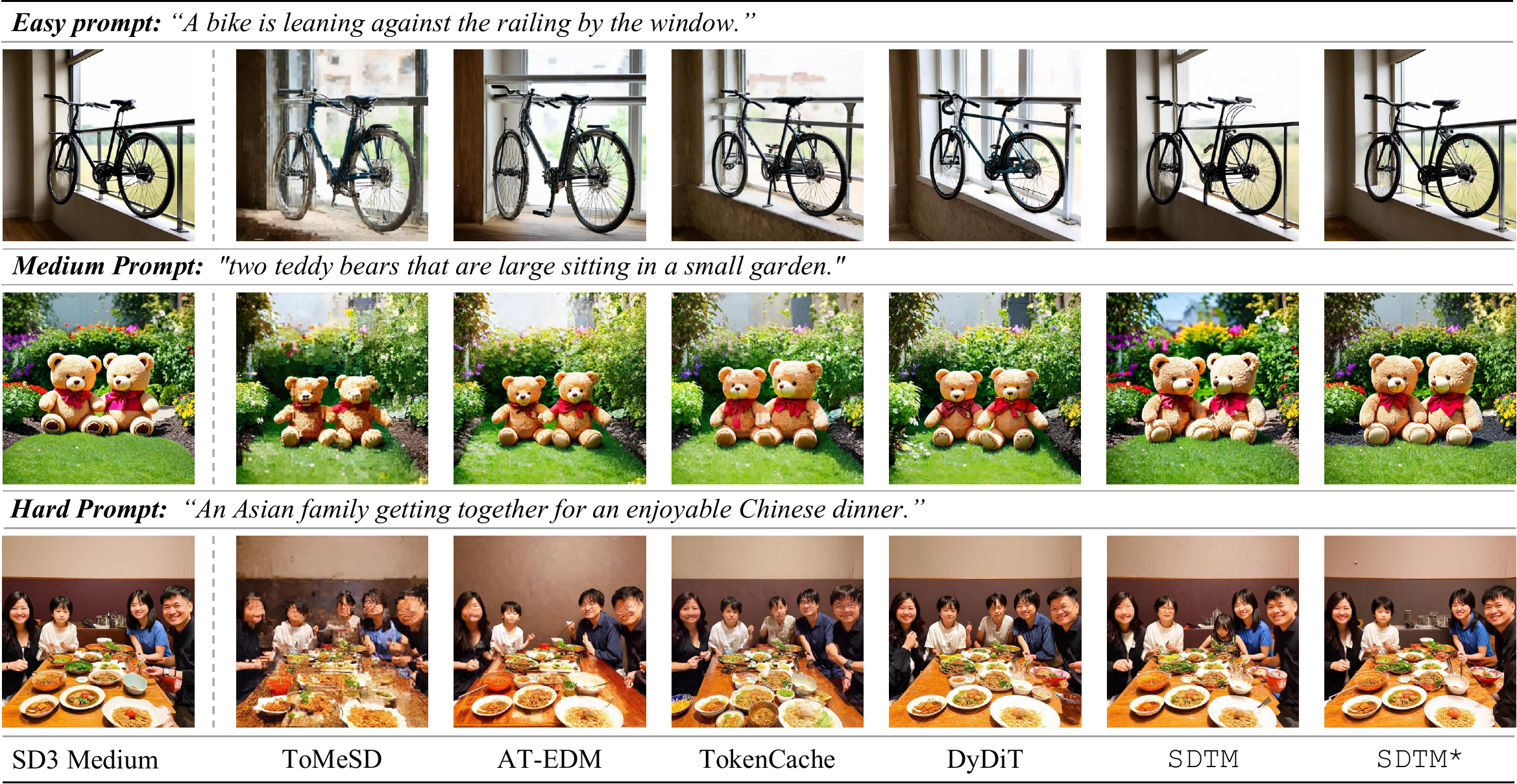}

	\caption{\textbf{Qualitative comparison} on COCO2017 and PartiPrompts under varying data complexities. For ToMeSD and AT-EDM, we use versions with approximately 1.3$\times$ acceleration, while others use approximately 1.5$\times$ versions. Best viewed when zoomed in.} 
    \vspace{-12 pt}
	\label{fig:comparison}
\end{figure*}

\subsection{Prompt Token Reweighting}
\label{sec:prompt}

The image generation process evolves from capturing the overall structure to refining intricate details, emphasizing the importance of the influence of prompt tokens at various stages~\cite{hertz2022prompt}. Moreover, as visual tokens become substantially compressed, the direction of prompt guidance grows increasingly vital. We developed prompt token reweighting to optimize the guiding direction across timesteps. We define the ``prompt'' as the image description, whereas ``instruction'' is the query to ChatGPT. As shown in Fig.~\ref{fig:overview}, given a prompt $\mathbb{P}$, we guide ChatGPT using the following interaction to categorize each prompt token $\mathbb{P}_k$.

\begin{tcolorbox}[title=Instruction: Categorize Prompt Token, boxrule=0pt, left=1mm, right=1mm, top=1mm, bottom=1mm, fontupper=\small]
    \textbf{System Instruction:} Suppose you are a data scientist. You will be provided with an [\emph{prompt}].
    You should categorize each prompt token into five categories: Strong Structure, Weak Structure, Neutral, Weak Detail, and Strong Detail. If a token contains both structure and detail, weight them for decision.\\
    \textbf{Context Instruction:} [\emph{prompt $\mathbb{P}$}]
\end{tcolorbox}
We multiply the attention values by an optimized range $[\alpha_p,\frac{1} {\alpha_p}]$. Take examples from Fig.~\ref{fig:overview}, the adjustment for weakly structured prompt token ``dog''  is $[\frac{\alpha_p}{2},\frac{2}{\alpha_p}]$, while for strongly detailed prompt token ``black'' is $[\frac{1}{\alpha_p}, \alpha_p]$.

\section{Experiment}

\subsection{Experiments Settings}

\noindent{\textbf{Implementation details.}} Our method can be seamlessly integrated into any text-to-image DiT architecture to facilitate post-training acceleration. It is available in two versions: \texttt{SDTM} is a straightforward implementation, and \texttt{SDTM}$^*$ adaptively adjusts the compression threshold based on the image's complexity. Unless specified otherwise, we set the initial $T-0.6T$ as the structure stage and the remaining $0.6T-0$ as the detail stage, using the hyperparameters basic ratio $\rho$ and maximum deviation $d$ to $0.5$ and $0.2$. 

\noindent{\textbf{Evaluations.}} We conduct extensive quantitative and qualitative experiments on various model configurations, including SD3 Medium, SD3.5 Large and SD3.5 Large Turbo~\cite{mmdit}, utilizing different schedulers such as Rectified Flow~\cite{rao2021dynamicvit}, DPM-Solver++~\cite{lu2022dpm++} across varying denoising steps (e.g. 50, 28, 20, 15). 
Following the protocol in AT-EDM~\cite{ATEDM}, our experiments were executed primarily on the COCO2017 validation set~\cite{mscoco} and PartiPrompt~\cite{parti} at an image resolution of $1024 \times 1024$. 
We evaluated performance using MACs and Latency, alongside FID~\cite{FID} and CLIP scores~\cite{clip} for image quality assessment. Latency was calculated by the average time required to generate $5000$ images on COCO2017 validation. All experiments were performed using $4$ NVIDIA A100 40G GPUs. 

\begin{figure}[t] 
    \centering
	\includegraphics[width=0.95\linewidth]{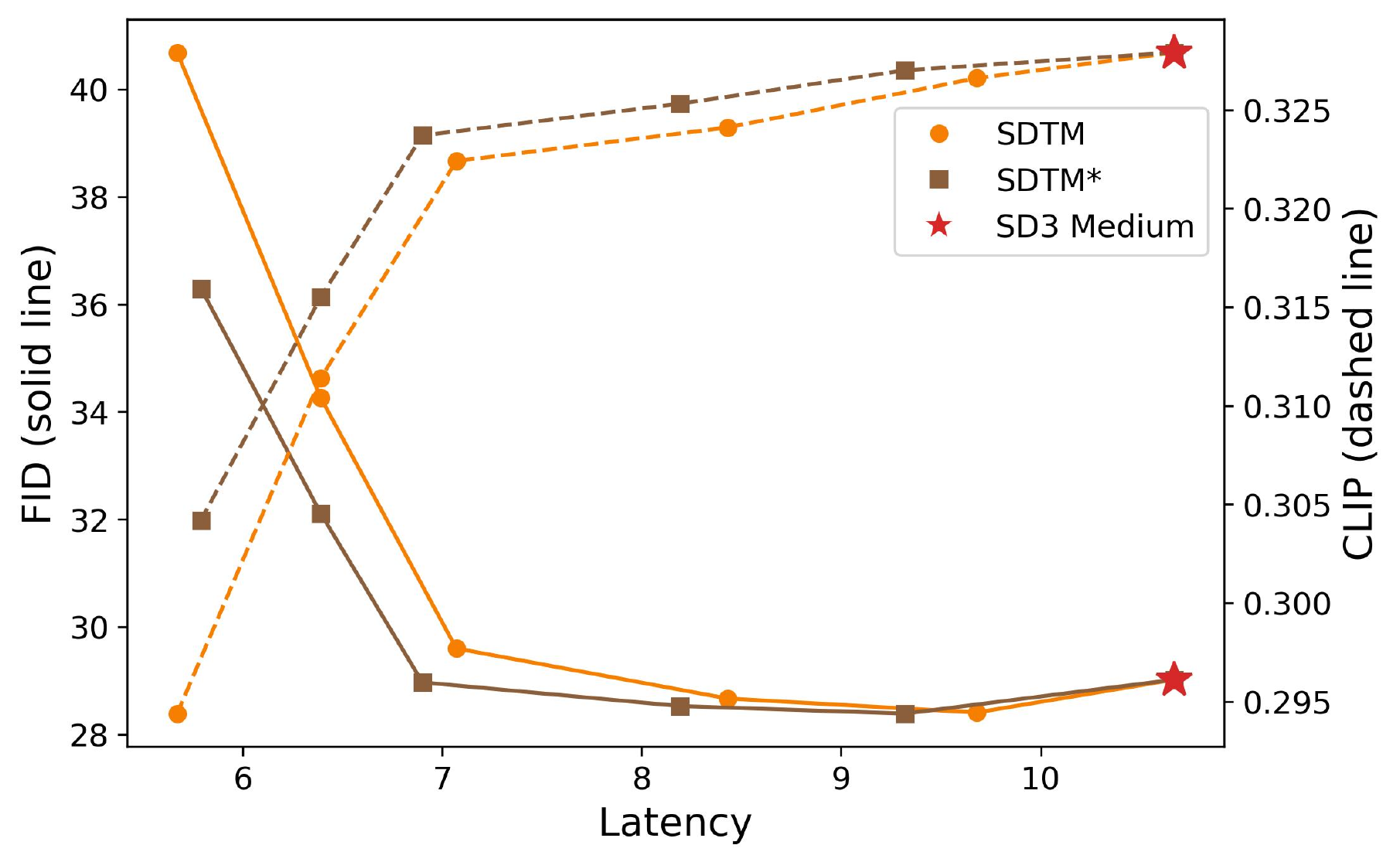}
        \vspace{-5 pt}
        \caption{The trade-off of Latency vs. FID and CLIP. We obtain \texttt{SDTM} and \texttt{SDTM*} with different latency by adjusting the ratio.}
        \vspace{-10 pt}
	\label{fig:tradeoff}
\end{figure} 

\subsection{Comparisons with SOTAs}

Table~\ref{table:sd3_comparison} compares our method with existing token-wise feature redundancy compression techniques. 
In configuration a, ToMeSD and AT-EDM show noticeable quality degradation, whereas our \texttt{SDTM} and \texttt{SDTM}* slightly improve image quality with reductions in FID by 0.29 and 0.45. 
In configuration b, as ToMeSD and AT-EDM exhibit significant quality declines, we introduce comparisons with finetuning-based TokenCache and DyDiT. 
Our approach achieves comparable image quality without the need for fine-tuning, benefiting from our detailed analysis and targeted reduction of feature redundancy at various stages. 
Our method achieves optimal CLIP, leveraging the prompt reweighting strategy that maintains directional guidance with fewer compressed tokens, which are detailed in ablation. 
Qualitative comparisons in Fig.~\ref{fig:comparison} further support our findings.
Moreover, \texttt{SDTM}* adaptively adjusts the merging threshold for samples of varying difficulty, allocating limited computational resources more effectively and surpassing other methods in image quality and alignment with the original images.

\subsection{Comparisons with baselines}

\begin{figure}[t] 
    \centering
	\includegraphics[width=1\linewidth]{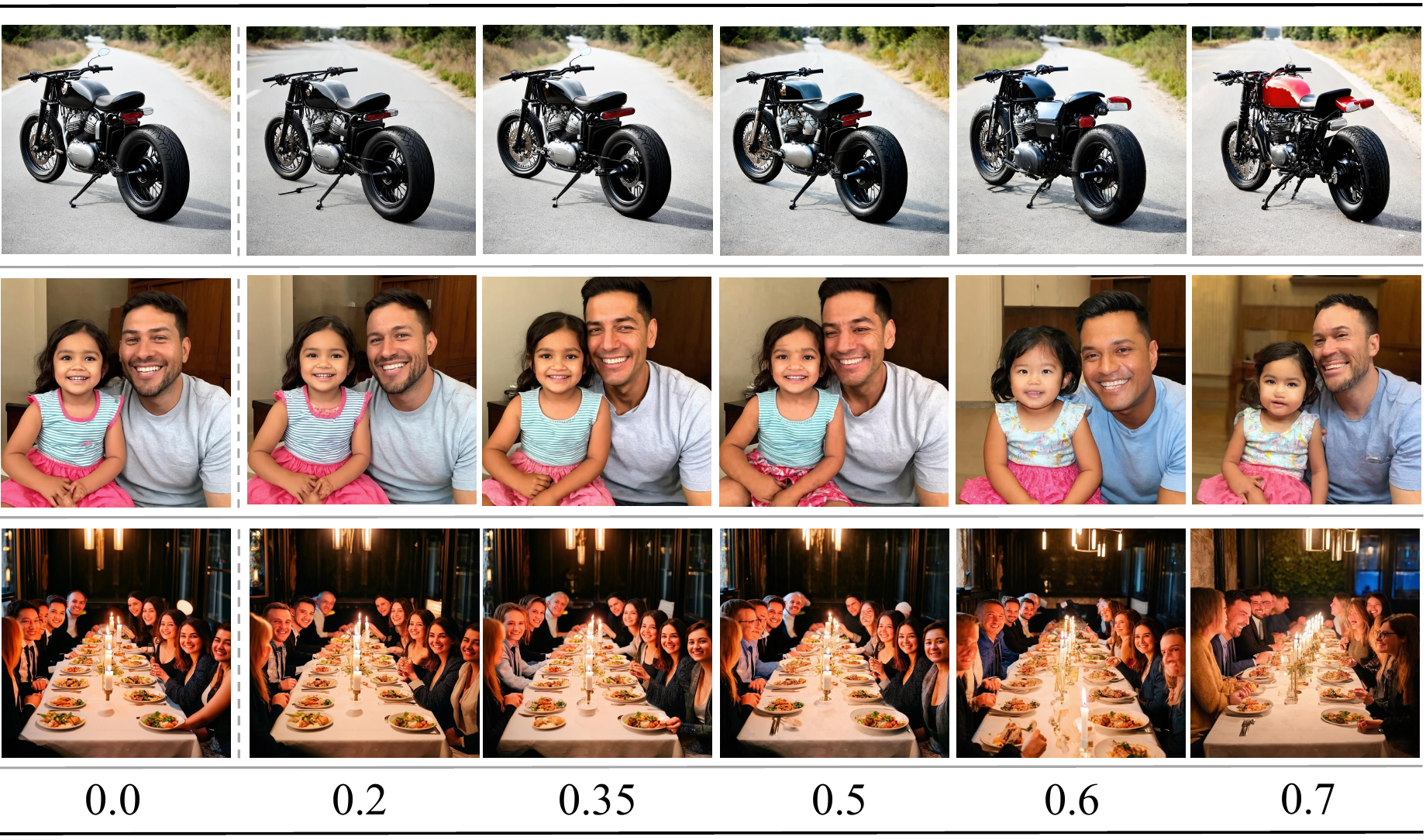}
        \vspace{-15 pt}
	\caption{Visualization of results for graduated basic ratios. 
    } 
	\label{fig:ratio}
\end{figure}

We further integrate \texttt{SDTM} and \texttt{SDTM}* into more baselines to evaluate their compatibility. As indicated in Fig.~\ref{fig:tradeoff} and Fig.~\ref{fig:ratio}, we evaluate the SD3 Medium integrated with our method across a range of basic ratios $\rho$ with the optimal deviation $d$ set at 0.2. Results show that image quality remains stable with 1.55$\times$ acceleration (config b). However, further increasing the acceleration ratio progressively degrades image quality, confirming that config b is the most advantageous balance. This phenomenon is attributed to the distribution of redundancy in the images: compression ratios that are too low fail to harness token merging techniques fully. 
In contrast, excessively high ratios cause a rapid accumulation of merging errors. Additionally, as shown in Table~\ref{table:baseline_all}, we extend \texttt{SDTM} and \texttt{SDTM}* to SD3 Medium and SD3.5 Large to explore the impact of different denoising steps (28, 20, and 15) and various schedulers (RF, DPM-Solver++). Our method shows high adaptability across diverse baselines and schedulers. \textit{More comparisons are available in the supplementary material. }

\begin{table}[t]
\centering
\resizebox{\linewidth}{!}{
    \setlength{\tabcolsep}{3pt}{
    \renewcommand{\arraystretch}{1}
    
    \begin{tabular}{lcccc}

        \toprule
        Method & Scheduler & W-MACs(T) $\downarrow$ & Latency(s) $\downarrow$ & FID $\downarrow$\\
        \midrule
        SD3 Medium & 28-RF & 168.4 & 6.29 & 28.74\\
        +\texttt{SDTM}&28-RF  & 103.3 & 4.22 & 28.95 \\
        +\texttt{SDTM}*&28-RF  & 101.5  & 4.11 & 28.67 \\
        \midrule
        SD3 Medium & 20-RF & 120.3 & 4.68 & 28.86\\
        +\texttt{SDTM}&20-RF  & 74.4 & 3.15 & 29.10 \\
        +\texttt{SDTM}*&20-RF  & 72.5 & 3.08 & 28.74 \\
        \midrule
        SD3 Medium & 20-DPM & 120.3 & 4.68 & 29.04\\
        +\texttt{SDTM}&20-DPM  & 74.4 & 3.16 & 29.31 \\
        +\texttt{SDTM}*&20-DPM  & 72.9 & 3.12 & 29.08 \\  
        \midrule
        SD3 Medium & 15-RF & 90.2 & 3.61 & 29.36\\
        +\texttt{SDTM}&15-RF  & 56.3 & 2.45 & 29.82 \\
        +\texttt{SDTM}*&15-RF  & 54.4 & 2.43 & 29.51 \\
        \midrule  
        SD3.5 Large & 28-RF & 563.5 & 18.54 & 25.91 \\
        + \texttt{SDTM}& 28-RF & 346.5 & 12.10 & 25.85 \\
        + \texttt{SDTM}*& 28-RF & 339.3 & 11.94 & 25.72 \\
        \bottomrule
    \end{tabular}}
}
\vspace{-5pt}
\caption{Comparison of our \texttt{SDTM} and \texttt{SDTM} approaches with SD3 Medium and SD3 Large across different schedulers. Here, “W-MACs” represent the total computation across all steps.*}
\vspace{-10pt}
\label{table:baseline_all}
\end{table} 

\subsection{Ablation studies}

We conduct ablation studies on the primary components of our method as outlined below. \textit{More detailed experiments and analyses are available in the supplementary material.}

\begin{figure}[t] 
    \centering
	\includegraphics[width=1\linewidth]{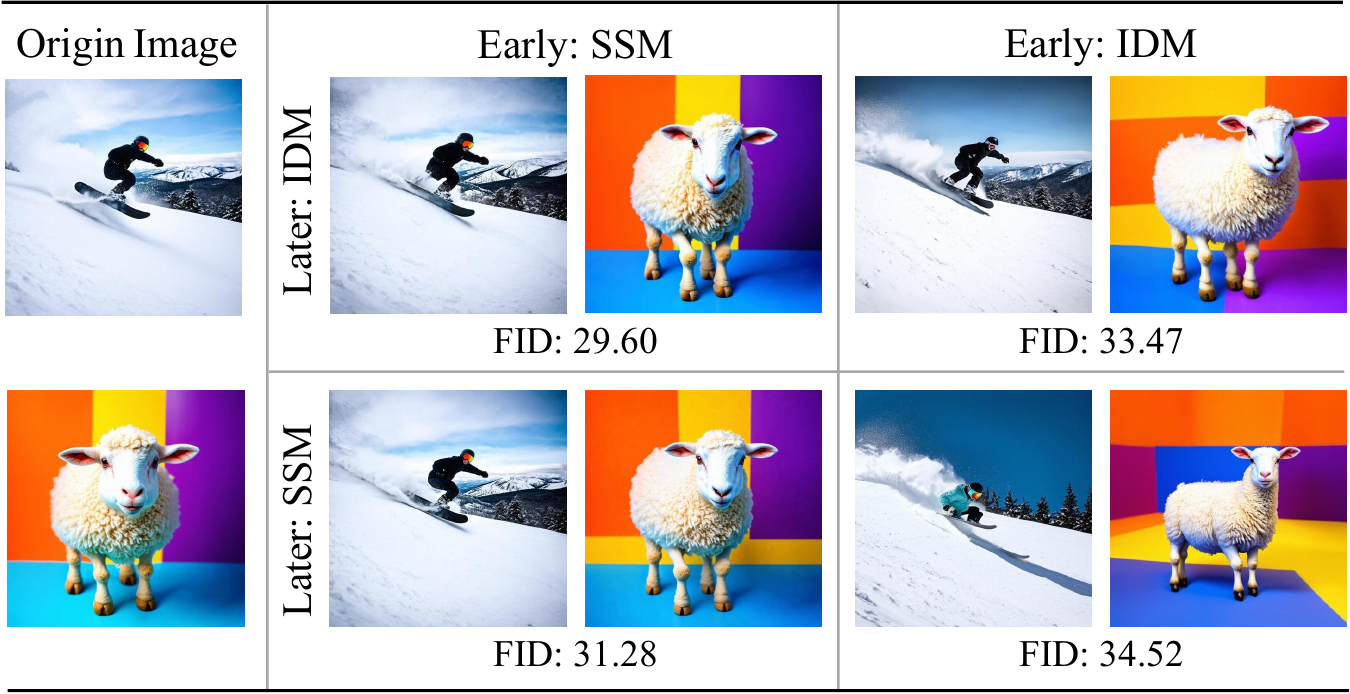}
	\vspace{-15 pt}
    \caption{Ablation study of different token merging combinations at various stages. SSM in early stages and IDM in later stages performs the best, while the reverse config performs the worst.} 
    \vspace{-5 pt}
	\label{fig:ablation_merge}
\end{figure}

\noindent \textbf{Effect of token merging strategies.}
To validate the effectiveness of our proposed structure-then-detail token merging strategies, we performed ablation studies using different combinations of merging strategies, as illustrated in Fig.~\ref{fig:ablation_merge}. The results indicate that employing SSM in the early stages and IDM in the later stages yields optimal performance, evidenced by an FID score of 29.60. In contrast, the IDM-then-SSM sequence recorded the poorest results, with an FID score of 34.52. These results align with our previous analysis in Sec.~\ref{sec:pre}, which posits that feature redundancy predominantly occurs among locally less-structure tokens in early stages and globally less-detail tokens in later stages. 

\begin{table}[t!]

	\centering
	\resizebox{1\linewidth}{!}{
    \setlength{\tabcolsep}{11pt}{
		\begin{tabular}{lcccc}
			\toprule
			Method       &deviation &decline & FID & CLIP\\
			\midrule
			\multirow{5}{*}{\rotatebox[origin=c]{0}{\texttt{SDTM}}}             
                &0.1  &cosine &30.53 & 0.3216 \\    
                &0.2  &linear  &31.64 & 0.3202  \\
			&0.2  &cosine  &29.60 & 0.3224    \\
   			&0.3  &cosine  &31.81 & 0.3198  \\
   			&0.4  &cosine  &36.03 & 0.3187  \\     
			\bottomrule
	\end{tabular}}}
 	\caption{Ablation of deviation values and ratio decline strategies. When the deviation of 0.2 and decline of cosine, the best results are achieved. We mark the optimal trade-off setting by $^\dagger$.}
    \label{tab:ablation_prune}
        \vspace{-15 pt}
\end{table}

\begin{figure}[t] 
    \centering
	\includegraphics[width=1\linewidth]{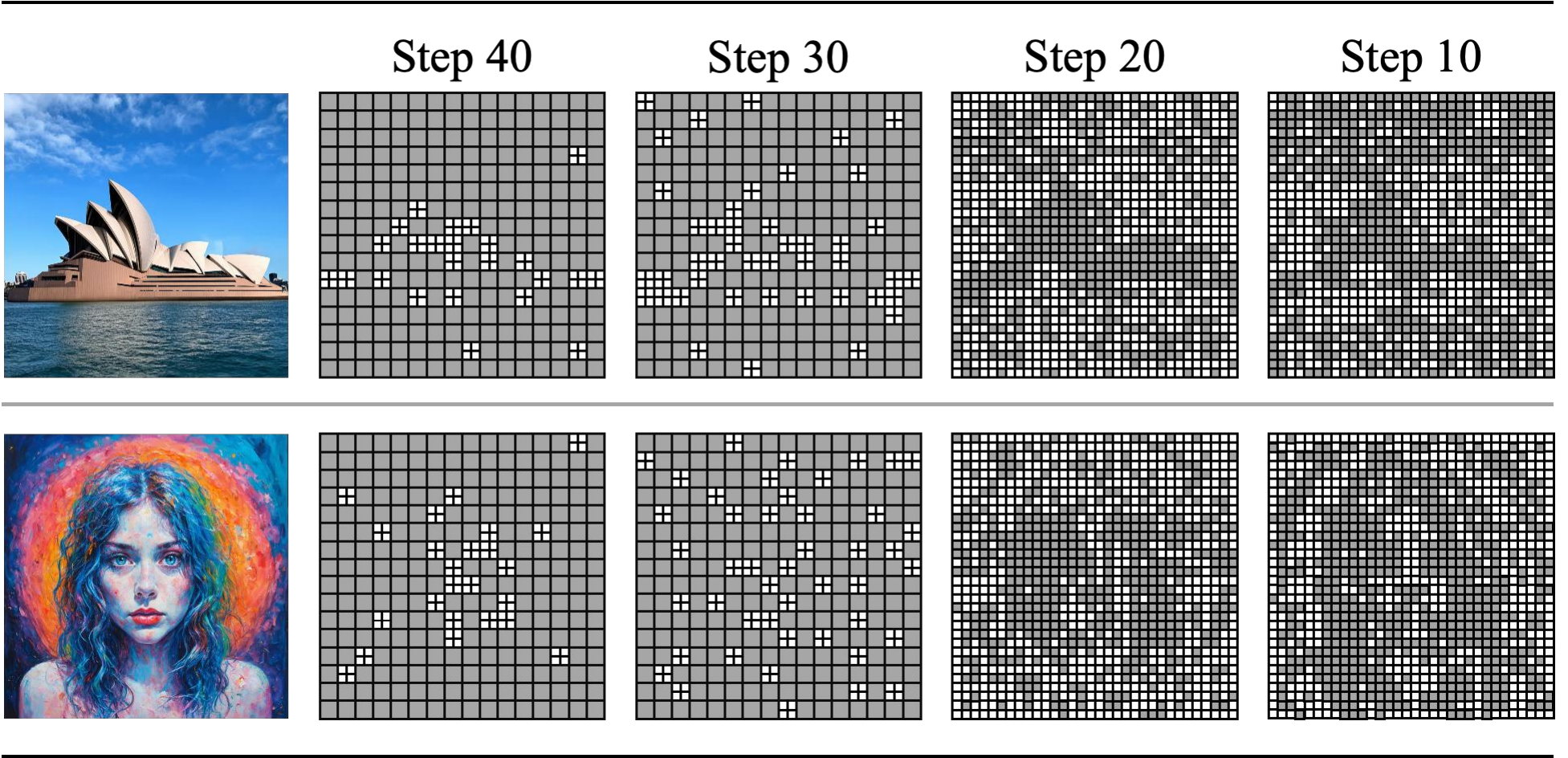}
        \vspace{-15 pt}
	\caption{Visualization of merged tokens selected. White masks represent independent sampling, while gray masks represent merging. In the later stage, only inattentive merged tokens are grayed.} 
        \vspace{-10 pt}
	\label{fig:show}
\end{figure}

\noindent \textbf{Effect of compression ratio adjusting.} Due to varying degrees of feature redundancy across denoising stages, we developed dynamic ratio adjusting and adaptive threshold adjusting strategies. Notably, the adaptive threshold adjustment mechanism distinguishes \texttt{SDTM*} from \texttt{SDTM}, as demonstrated by its ability to dynamically optimize merging ratios for samples with diverse complexities (as shown in Fig.~\ref{fig:tradeoff} and Table~\ref{table:baseline_all}). We conducted ablation on the deviation value $d$ and ratio decline strategies within the dynamic ratio adjusting framework in Table~\ref{tab:ablation_prune}. Our results indicate that a deviation of $d = 0.2$ achieves the optimal trade-off, while cosine decay yields superior performance compared to linear decay. Furthermore, Fig.~\ref{fig:show} illustrates the progressive merging process within SSM and IDM merging strategies throughout the generation process. 

\begin{figure}[t] 
    \centering
	\includegraphics[width=1\linewidth]{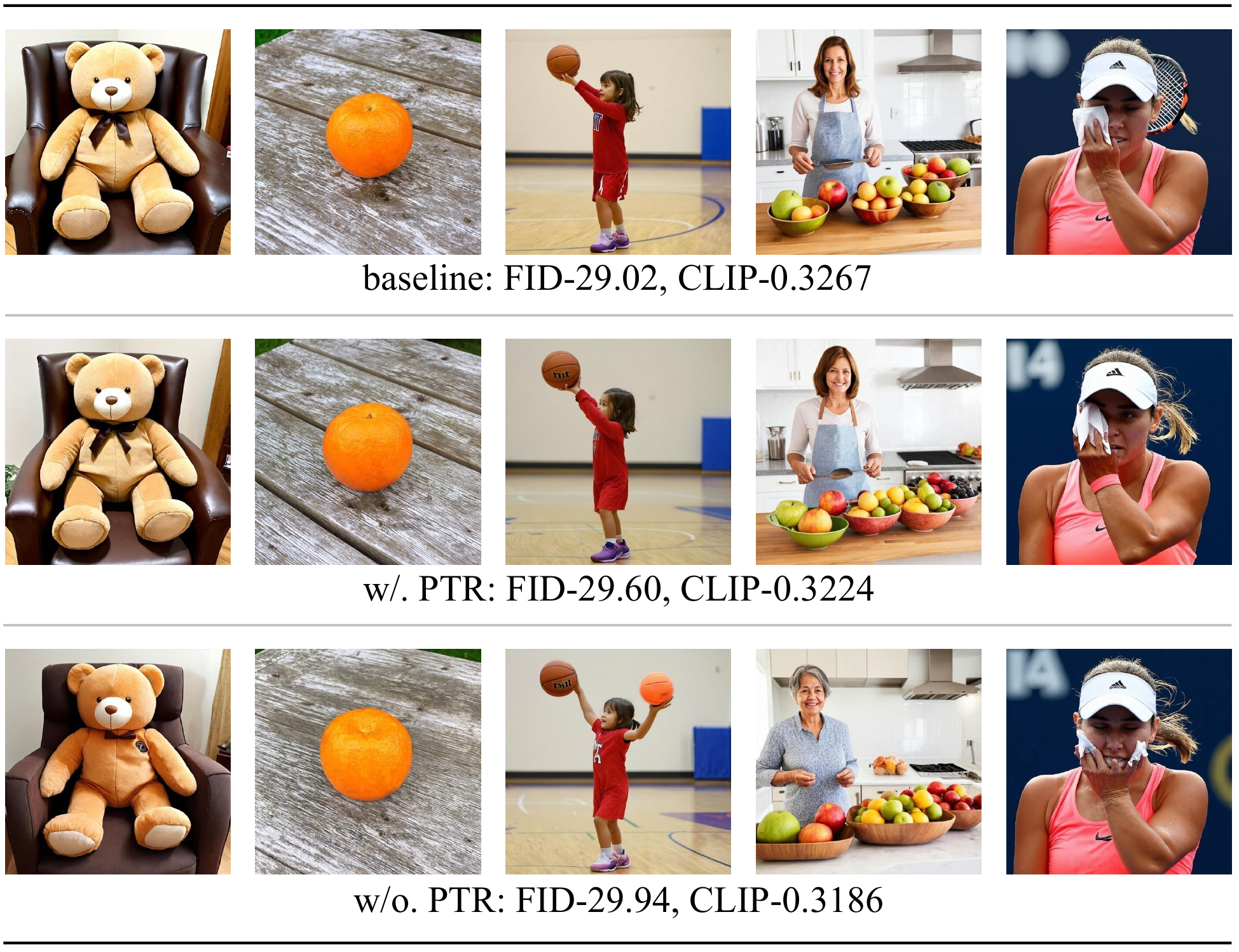}
        \vspace{-15 pt}
	\caption{Ablation of prompt token reweighting (PTR). From left to right, the absence of PTR leads to minor color, structural errors, and severe semantic misunderstandings.} 
	\label{fig:ablation_ptr}
        \vspace{-15 pt}
\end{figure}

\noindent \textbf{Effect of prompt token reweighting.} PTR optimizes the guidance direction at various stages. Fig.~\ref{fig:ablation_ptr} shows that employing PTR slightly enhances the FID and significantly improves CLIP. We address a prevalent issue identified in ToMeSD and AT-EDM: token compression undermines CLIP. We attribute this to reduced attention to images due to the fewer tokens caused by compression. Fig.~\ref{fig:ablation_ptr} also shows that the absence of PTR leads to color, structural errors, and semantic misunderstandings. Therefore, while PTR does not directly reduce computation costs, it is crucial to maintain alignment with prompts.


\section{Conclusion}

In this paper, we conduct a detailed analysis of the location and degree of feature redundancies and design a novel approach to accelerate DiTs by targeting redundancies in areas overlooked by the denoising process. Our innovative \texttt{SDTM} method dynamically addresses less-structure and less-detail redundancy throughout the generation process. It can be integrated seamlessly into any existing DiT architecture, accelerating generation without additional fine-tuning. We conducted extensive quantitative and qualitative experiments to demonstrate the effectiveness of our method across various architectures and schedulers. \textbf{Limitation:} Due to the inherent trade-offs of compression ratios in token merging, greater compression demands necessitate the integration of other acceleration techniques, such as distillation.

{
    \small
    \bibliographystyle{ieeenat_fullname}
    \bibliography{main}
}
\clearpage
\setcounter{page}{1}
\maketitlesupplementary

\noindent\textbf{We organize our supplementary material as follows.} 
\vspace{-12 pt}
\paragraph{Implementation Details:}
\begin{itemize}
    \item Sec.~\ref{app_sec:pre}: Details of preliminary experiments.
    \item Sec.~\ref{app_sec:sotas}: Details of the implementation of backbones, our methodologies, and SoTA methods.
\end{itemize} %
\noindent\textbf{Additional Analyses:}
\begin{itemize}
    \item Sec.~\ref{app_sec:reducedmacs}: Analysis of reduced MACs across timesteps.
    \item Sec.~\ref{app_sec:merge_cost}: Analysis of computation overhead of SSM and IDM merging mechanisms.
\end{itemize} %
\noindent\textbf{Additional Experiments:}
\begin{itemize}
    \item Sec.~\ref{app_sec:baseline}: Extended comparisons with baseline models.
    \item Sec.~\ref{app_sec:imagesize}: Ablation study on image size effects.
    \item Sec.~\ref{app_sec:qualitative}: More visualizations of generated images.
\end{itemize} %

\section{Implementation Details}
\subsection{Details of Preliminaries}
\label{app_sec:pre}

In Sec.~3, we examine the evolution of the LL, HL, LH, and HH subbands in the denoising process, along with the location and degree evolution of feature redundancies. Below, we detail the preliminary experiments.

\vspace{-10pt}
\paragraph{Feature Collection.} Utilizing the baseline model of Stable Diffusion 3 Medium and a 50-step Rectified Flow scheduler, we investigated the estimated noise $x_k^{(t,l)}$ generated by each transformer block at every step across 10,000 samples. Here, $t$ represents the step, $l$ denotes the block index, and $k$ indicates the sample index. These samples included 5,000 prompts each from the MS-COCO 2014 validation split and MS-COCO 2017 validation split. 

\vspace{-10pt}
\paragraph{Evolution of Denoising Process.} 
Based on multi-wavelet functions, the discrete wavelet transform (DWT) decomposes an input into four wavelet coefficients: LL, LH, HL, and HH. The LL coefficient captures the low-frequency component, reflecting structural features, while LH, HL, and HH coefficients detect high-frequency components associated with detail features. Initially, we apply DWT to each $x_k^{(t,l)}$ for decomposition into $x_k^{(t,l)} = \{(x_{k, LL}^{(t,l)}, x_{k, LH}^{(t,l)}, x_{k, HL}^{(t,l)}, x_{k, HH}^{(t,l)})\}$. Subsequently, we compute the L2 normalization for each subband of $x_k^{(t,l)}$. Finally, we calculate the maximum, minimum, and average values across batch and layer dimensions to derive $\|x^{(t)}\|_2$, which are used to generate the Figs.~2~(a) and~(b).

\vspace{-10pt}
\paragraph{Evolution of Feature Redundancies.} Initially, we compute the cosine similarity for each token pair in $x_k^{(t,l)}$. Using the output similarity matrix, we identify the most similar token $x_{k,(i’,j’)}^{(t,l)}$ for every token $x_{k,(i,j)}^{(t,l)}$, where $(i, j)$ and $(i’, j’)$ are coordinates of a token and its most similar counterpart. We assess the location and degree of token redundancies by calculating the mean L2 distance and the cosine similarity among the closest pairs. In Fig.~2~(c), we calculate the L2 distance for each token pair $(i, j)$ and $(i’, j’)$, averaging these distances across pairs and batches to compute the mean L2 distance for each layer and step. We then average these results across the layer dimension to track the evolution across timesteps. Similarly, in Fig.~2~(d), we compute the cosine similarity for each token pair, averaging these values across pairs and batches to derive the mean similarity for each layer and step and averaging across the layer dimension to track the evolution along timesteps.

\subsection{Details of Backbones, Ours, and SoTAs}
\label{app_sec:sotas}
We implement our methods and other SoTA techniques, including ToMeSD, AT-EDM, TokenCache, and DyDiT, on the SD3 Medium model for evaluation.

\begin{figure}[t] 
    \centering
	\includegraphics[width=1\linewidth]{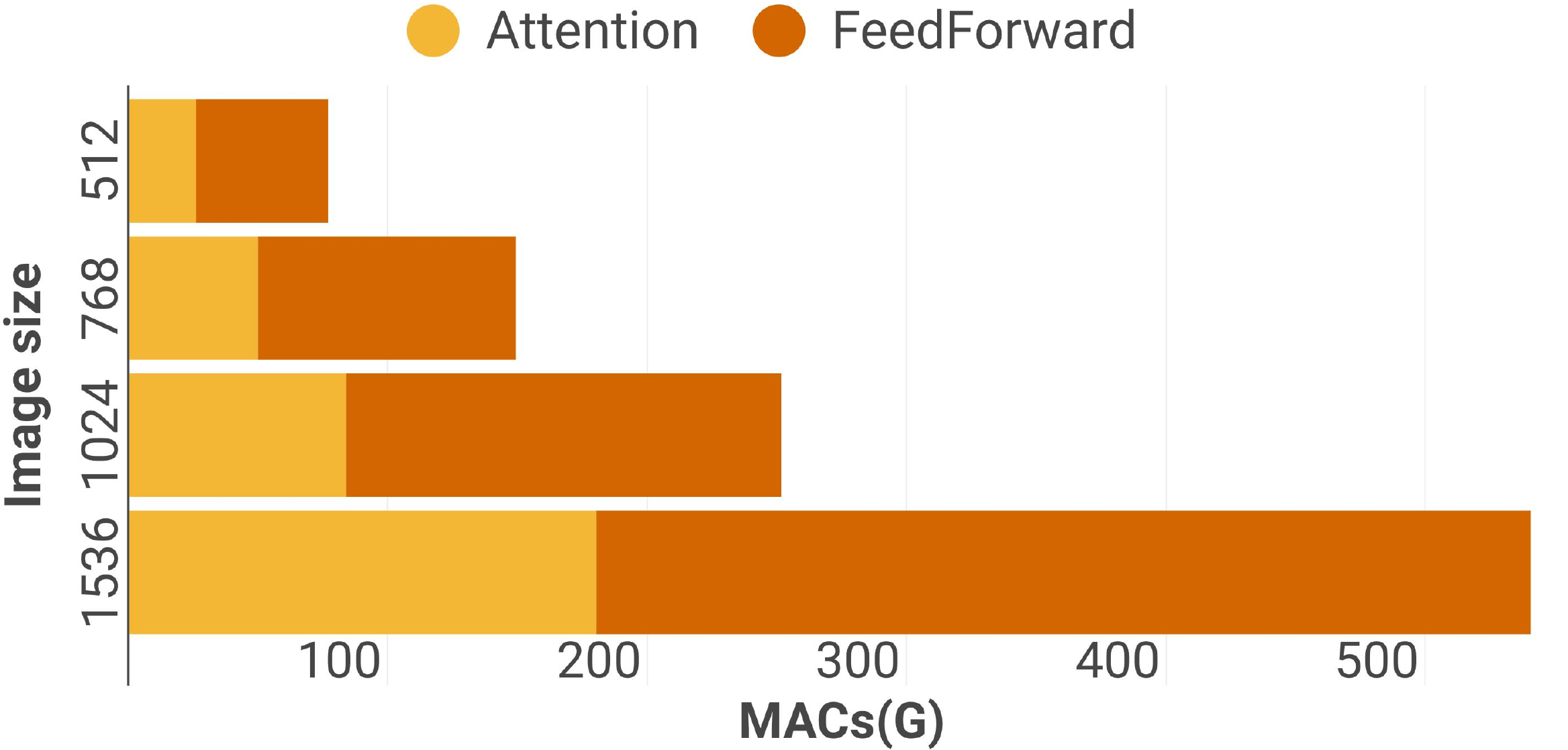}
	\caption{MACs of main components in different image sizes. 
    } 
        \vspace{-5pt}
	\label{fig:macs}
        \vspace{-5pt}
\end{figure}
\vspace{-10pt}
\paragraph{The SD3 Medium Backbone.} The SD3 Medium employs an advanced transformer architecture, the multimodal diffusion transformer (MMDiT). Distinct from diffusion models that rely on U-Net architectures, it incorporates 24 JointTransformerBlocks at uniform feature levels, enabling joint attention interactions between prompt and image tokens. The SD3 model employs the Rectified Flow scheduler with a default CFG of 7.0 and 50 / 28 denoising steps. 
Fig.~\ref{fig:macs} illustrates the MACs of each computational component within the JointTransformerBlocks. It is observed that the FeedForward accounts for approximately $2/3$ of the MACs, while the Attention mechanism occupies about $1/3$. In the following, we implemented the SoTAs token reduction methods for DiT to reduce the computation.

\vspace{-10pt}
\paragraph{Ours.} We implement our method in the SD3 Medium using a two-stage approach. In the early stage, we execute similarity-prioritized structure merging before the MHSA and MLP blocks. In the later stage, we perform inattentive-prioritized detail merging prior to the MLP blocks. Additionally, we dynamically adjust the compression ratio and prompt weights at each denoising step using compression ratio adjusting and prompt token reweighting. Depending on the settings of CRA, we differentia between dynamic ratio and adaptive threshold, resulting in two variants: \texttt{SDTM} and \texttt{SDTM}$^*$. During the batch inferences of \texttt{SDTM}$^*$, if the batch size exceeds one, inconsistencies in token numbers across the batch may occur due to adaptive threshold filtering. To mitigate this, we balance the difficulty of prompts within each batch using GPT scores and select the minimum number of tokens for pruning.

\vspace{-10pt}
\paragraph{ToMeSD.} ToMeSD employs a uniform merging strategy across all sampling steps, utilizing a 2D stride-based strategy to merge tokens. This strategy is applied on blocks at the highest-resolution feature level. Since all blocks in the MMDiT architecture share the same feature level, we initially experimented with applying ToMeSD across all Transformer blocks. However, this led to significant degradation in generation quality. Given that ToMeSD was originally designed for UNet-based DMs, where convolution and transformer modules alternate (and thus not all modules incorporate token merging), indiscriminate application of token merging to all Transformer modules in MMDiT proved suboptimal. To address this, we adapted the application of ToMeSD for MMDiT by implementing a staggered compression pattern: for every four consecutive Transformer blocks, we apply no compression, standard compression, reduced compression, and standard compression rates, respectively. This configuration achieve better generation quality in equivalent overall cost. Furthermore, we implemented ToMeSD's linearly interpolating of the ratio. 

\vspace{-10pt}
\paragraph{AT-EDM.} AT-EDM adopts a two-stage token pruning strategy, where $T-0.7T$ serves as the early stage and $0.7T-0$ as the later stage. This method uses a graph-based algorithm for token pruning across multiple cascaded attention block groups. We structured the MMDiT into six groups to align with this configuration. Additionally, AT-EDM incorporates a DSAP schedule that preserves unpruned attention blocks at the mid-stage due to the low feature level; this concern is not present in MMDiT. AT-EDM leaves the first attention block in each down-stage and the last in each up-stage unpruned. Consequently, we designate the first three groups as down-stage and the last three as up-stage to mirror this architecture.

\vspace{-10pt}
\paragraph{TokenCache.} TokenCache deploys a TPRR token pruning schedule, where $T-M$ serves as Phase I and $M-0$ as Phase II. During each phase, a cyclic schedule is employed, alternating between $I$-steps, with no pruning, and $K$-steps, where a Cache Predictor is trained to prune non-essential tokens. We implement TokenCache in SD3 Medium using this strategy and train the Cache Predictor according to the procedures outlined in their paper. We determine the optimal values of $(M, K_1, K_2)$, based on their ablation study, to be $(0.5T, 4, 2)$, where $K_1$ and $K_2$ correspond to different steps for Phase I and Phase II.

\vspace{-10pt}
\paragraph{DyDiT.} DyDiT introduces a timestep-wise dynamic width (TDW) to reduce model width and a spatialwise dynamic token (SDT) strategy to minimize redundancy at spatial locations. We integrated these strategies into SD3 Medium, using the FLOPs-aware end-to-end training method proposed by them to train the model. To ensure training stability, DyDiT incorporates fine-tuning stabilization, which is crucial for the FLOPs-aware end-to-end training method. Consequently, we conducted four training sessions and selected the optimal results. For the hyperparameter of cache interval, we chose an interval of 2, as identified as optimal in their ablation study.

\subsection{Details of Evaluations}
\label{app_sec:evaluation}
Our evaluation adheres to the settings employed in AT-EDM, conducting experiments on the COCO2017 validation split. We implemented a prompt deduplication strategy to ensure unique pairings of each image in the validation set with one prompt. Each image is center-cropped and resized for comparison. For metric calculations, we utilize the clean-fid\footnote{https://github.com/GaParmar/clean-fid/tree/main} to compute FID scores and the ViT-G/14 model from Open-CLIP\footnote{https://github.com/mlfoundations/open-clip} to calculate CLIP scores. Unless specified otherwise, all experiments were conducted using two NVIDIA A100 GPUs, generating images of 1024$\times$1024 resolution with a batch size of four.

\section{Additional Analyses}
\begin{figure}[t] 
    \centering
	\includegraphics[width=1\linewidth]{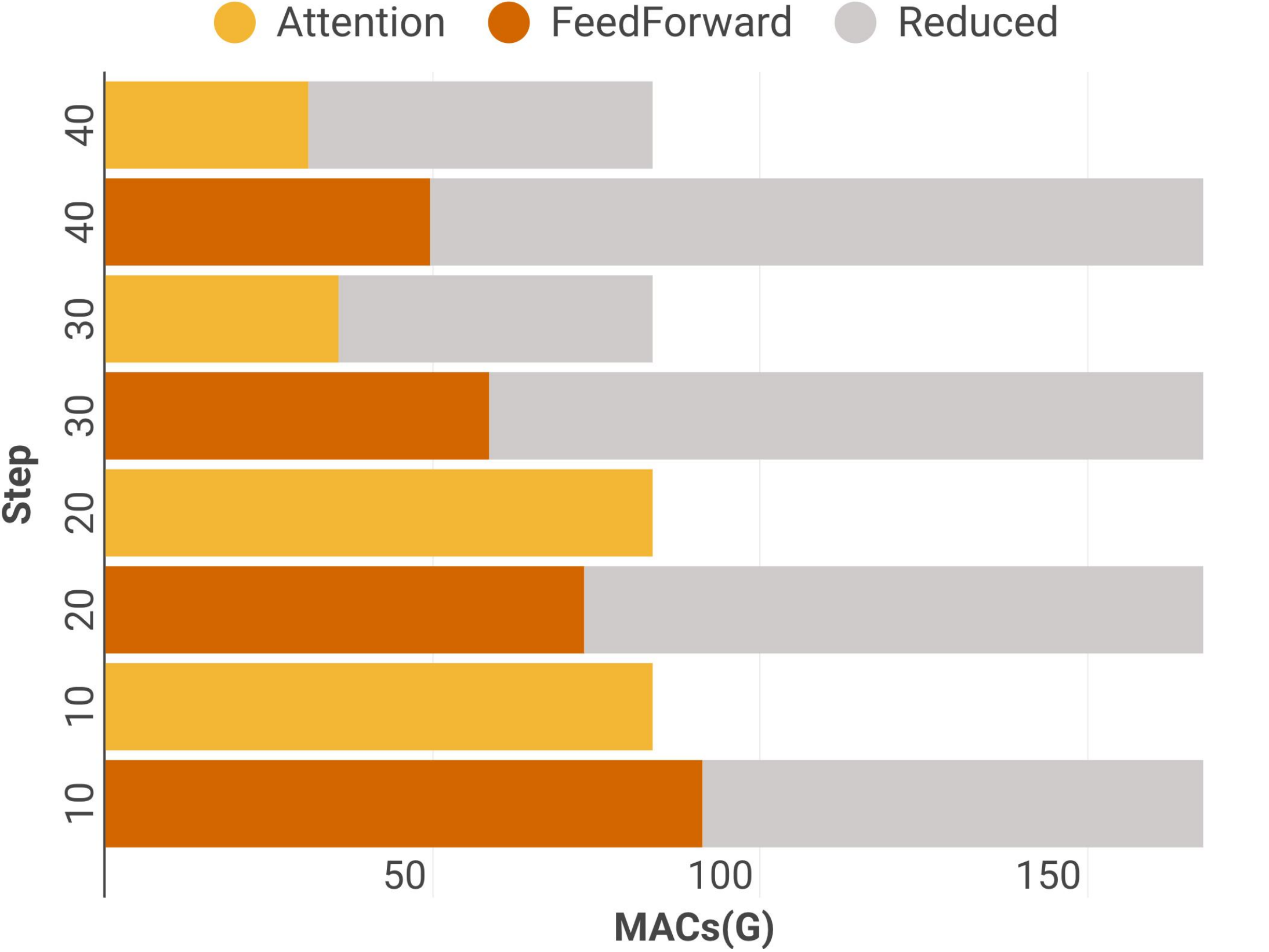}
	\caption{Reduced MACs of main components along timesteps. 
    } 
                \vspace{-5pt}
	\label{fig:macsave}
            \vspace{-5pt}
\end{figure}

\subsection{Reduced Computation via Timesteps}
\label{app_sec:reducedmacs}
In our methodology, we introduce similarity-prioritized structure merging to enhance the efficiency of MHSA and MLP blocks during the structure stage, and inattentive-prioritized detail merging to speed up MLP blocks in the detail stage. Additionally, we progressively decrease the compression ratio as denoising progresses. Based on the SD3 Medium model with our \texttt{SDTM} approach, we generate images with a resolution of 1024$\times$1024 and display the MACs of MHSA and MLP blocks at representative steps 40, 30, 20, and 10 in Fig.~\ref{fig:macsave}. The results indicate increasing computational allocation across the timesteps, aligning with the evolution of feature redundancies over time as discussed in Sec.~3.

\subsection{Identification Computation}
\label{app_sec:merge_cost}

Although similarity-prioritized structure merging and inattentive-prioritized detail merging reduce the computation of original transformer blocks, they introduce additional cost for identifying similarity and inattentive redundancy. Assuming the feature \(X \in \mathbb{R}^{N, D}\) and a window size \(m \times m\), we report the complexity and MACs for redundancy identification in Table~\ref{tab:rebuttal_cost}. Results show that for both the SD3 Medium and SD3.5 Large models, the computational cost of redundancy identification is negligible compared to the overall model cost.
\begin{table}[!t]
\centering
\resizebox{\linewidth}{!}{
\setlength{\tabcolsep}{0.5mm}{
\begin{tabular}{lccc}
    \toprule
    \textbf{Object}
     & Complexity & SD3-M MACs  & SD3-L MACs \\
    \midrule
    Model & - & 6.01T & 20.13T \\
    Similarity & $\mathcal{O}(ND + \frac{N}{m^2} \log(\frac{N}{m^2}))$ & 3.0E-04T & 1.0E-03T \\
    Inattentive & $\mathcal{O}(N^2 + N \log N)$ & 8.1E-04T & 2.7E-03T \\
    \bottomrule
\end{tabular}
}}
\vspace{-5pt}
\caption{Complexity and cost analysis of identifying redundancy.}
\label{tab:rebuttal_cost}
\vspace{-10pt}
\end{table}

\section{Additional Results}
\label{sec:moreresults}

\subsection{More Comparisons with Baselines}
\label{app_sec:baseline}

\noindent\textbf{More Comparisons with Baselines.} We expanded the integration of \texttt{SDTM} and \texttt{SDTM}* into additional baselines to assess their adaptability. These included the SD3.5 Large Turbo, a distilled version of SD3.5 Large designed to enhance image quality with fewer denoising steps\footnote{https://stability.ai/news/introducing-stable-diffusion-3-5}; and FLUX.1-dev, a 12 billion-parameter rectified flow transformer noted for its advanced performance in image generation. We utilized their default CFG values and recommended schedulers. For FLUX.1-dev, we observed that including Rotary Positional Embedding in the MHSA blocks is extremely sensitive to token reduction; therefore, we left these MHSA blocks unpruned.  The outcomes, presented in Table~\ref{table:baseline_more}, indicate that despite considerable reductions in steps and the constraints imposed on MHSA, which lower the compression ratio, our method still achieves a favorable acceleration while maintaining the image quality.

\begin{table}[t]
\centering
\resizebox{\linewidth}{!}{
    \setlength{\tabcolsep}{3pt}{
    \renewcommand{\arraystretch}{1}
    
    \begin{tabular}{lcccc}

        \toprule
        Method & Step & W-MACs(T) $\downarrow$ & Latency(s) $\downarrow$ & FID $\downarrow$\\
        \midrule
        SD3.5 Large Turbo & 4 & 47.7 &0.69 & 30.48\\
        +\texttt{SDTM}&4  & 33.4 & 0.53 & 31.25 \\
        +\texttt{SDTM}*&4  & 32.8 & 0.52 & 30.62 \\
        \midrule
        SD3.5 Large Turbo & 10 & 119.4 & 1.30 & 30.27\\
        +\texttt{SDTM}&10  & 75.6 & 0.89 & 31.01 \\
        +\texttt{SDTM}*&10  & 74.5 & 0.87 & 30.41 \\ 
        \midrule
        FLUX.1-dev & 20 & 595.2 & 13.81 & 30.25\\
        +\texttt{SDTM}$^\dagger$ &20 & 386.5 & 10.15 & 30.68 \\
        +\texttt{SDTM}*$^\dagger$ &20 & 381.4 & 10.08 & 30.21 \\ 
        \bottomrule
    \end{tabular}}
}
\vspace{-5pt}
\caption{Comparisons of our \texttt{SDTM} and \texttt{SDTM} with SD3 Large Turbo and FLUX.1-dev at various steps. A dagger $\dagger$ indicates that the MHSA block is not accelerated for adaptation.}
\vspace{-10pt}
\label{table:baseline_more}
\end{table}

\subsection{Ablation of Image Size} 
\label{app_sec:imagesize}
Following the settings of AT-EDM, our experiments were primarily conducted on 1024 px images. To assess our method’s applicability across different image sizes, we tested resolutions including 512, 768, and 1536 px, with results detailed in Table~\ref{table:image_size}. We utilized the SD3 Medium equipped with a 50-RF scheduler as our baseline. It is important to note that the FID scores for different image sizes are not comparable since they correspond to distinct distributions. The results demonstrate that smaller image sizes slightly compromise image quality; this reduction can be attributed to the decreased similarity between patches at smaller sizes, which leads to less feature redundancy. Nonetheless, our approach consistently delivers significant acceleration compared to the baseline method.

\begin{table}[t]
\centering
\resizebox{\linewidth}{!}{
    \setlength{\tabcolsep}{5pt}{
    \renewcommand{\arraystretch}{1}
    
    \begin{tabular}{lcccc}

        \toprule
        Method & Image size & MACs(T) $\downarrow$ & Latency(s) $\downarrow$ & FID $\downarrow$\\
        \midrule
        baseline & 512 & 1.83 & 2.47 & 28.74\\
        +\texttt{SDTM}&512  &1.10   &1.61  & 29.86 \\
        \midrule
        baseline & 768 & 3.58 & 5.39 & 29.27\\
        +\texttt{SDTM}&768  &2.16  &3.54  &29.65  \\ 
        \midrule
        baseline & 1536 & 12.97 & 31.29 & 28.06\\
        +\texttt{SDTM} &1536  &8.04  &21.28  &28.24 \\ 
        \bottomrule
    \end{tabular}}
}
\vspace{-5pt}
\caption{Ablation of our \texttt{SDTM} with the SD3 Medium using a 50-RF scheduler across various image sizes.}
\vspace{-10pt}
\label{table:image_size}
\end{table}

\begin{figure*}[!t] 
    \centering
	\includegraphics[width=0.77\linewidth]{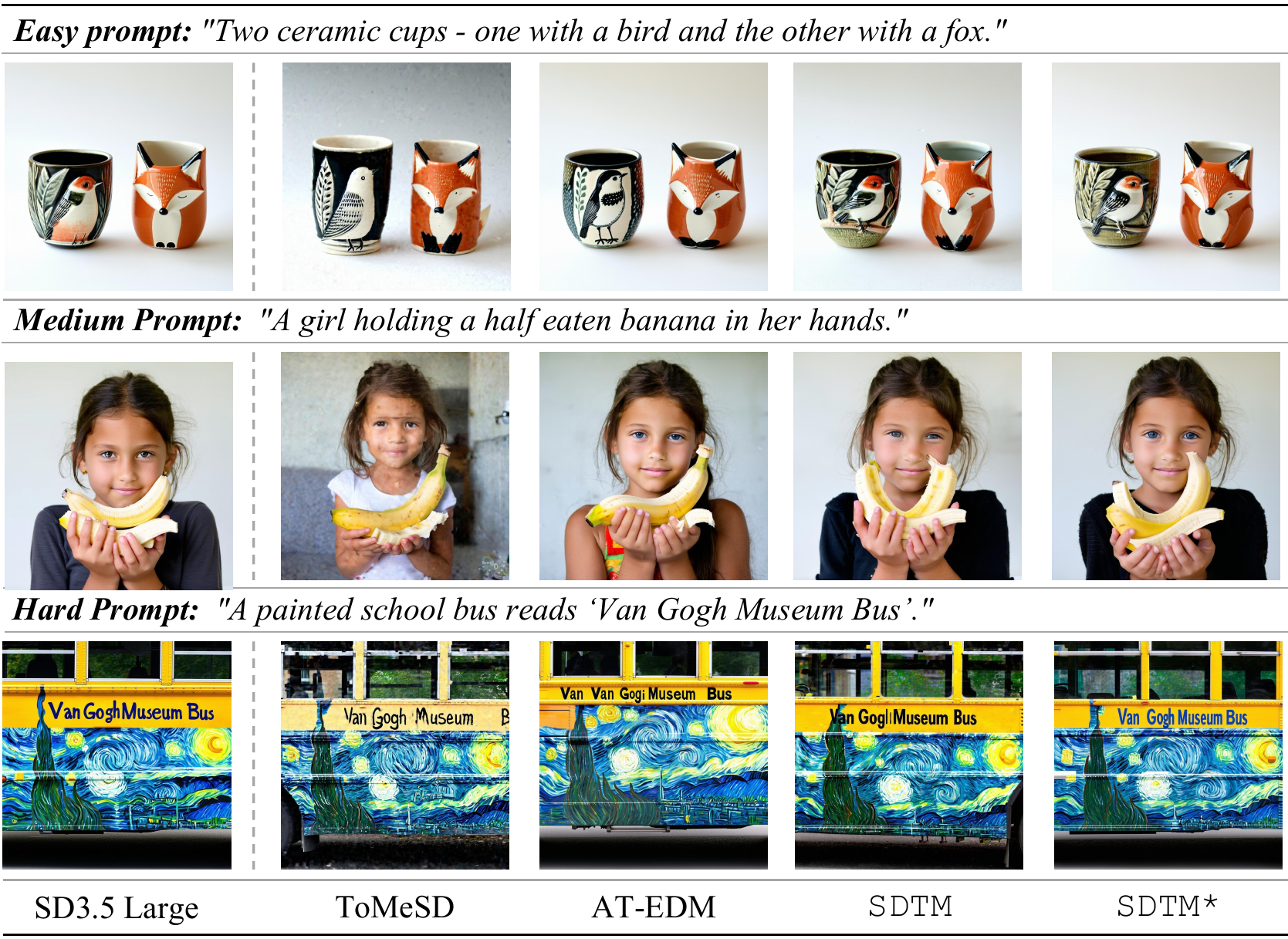}

	\caption{Qualitative comparison on SD3.5 Large under varying data complexities. For ToMeSD and AT-EDM, we use versions with approximately 1.3$\times$ acceleration, while others use approximately 1.5$\times$ versions. Best viewed when zoomed in.} 
    \vspace{-5 pt}
	\label{fig:comparison_supp}
\end{figure*}
\begin{figure*}[!t] 
    \centering
	\includegraphics[width=0.82\linewidth]{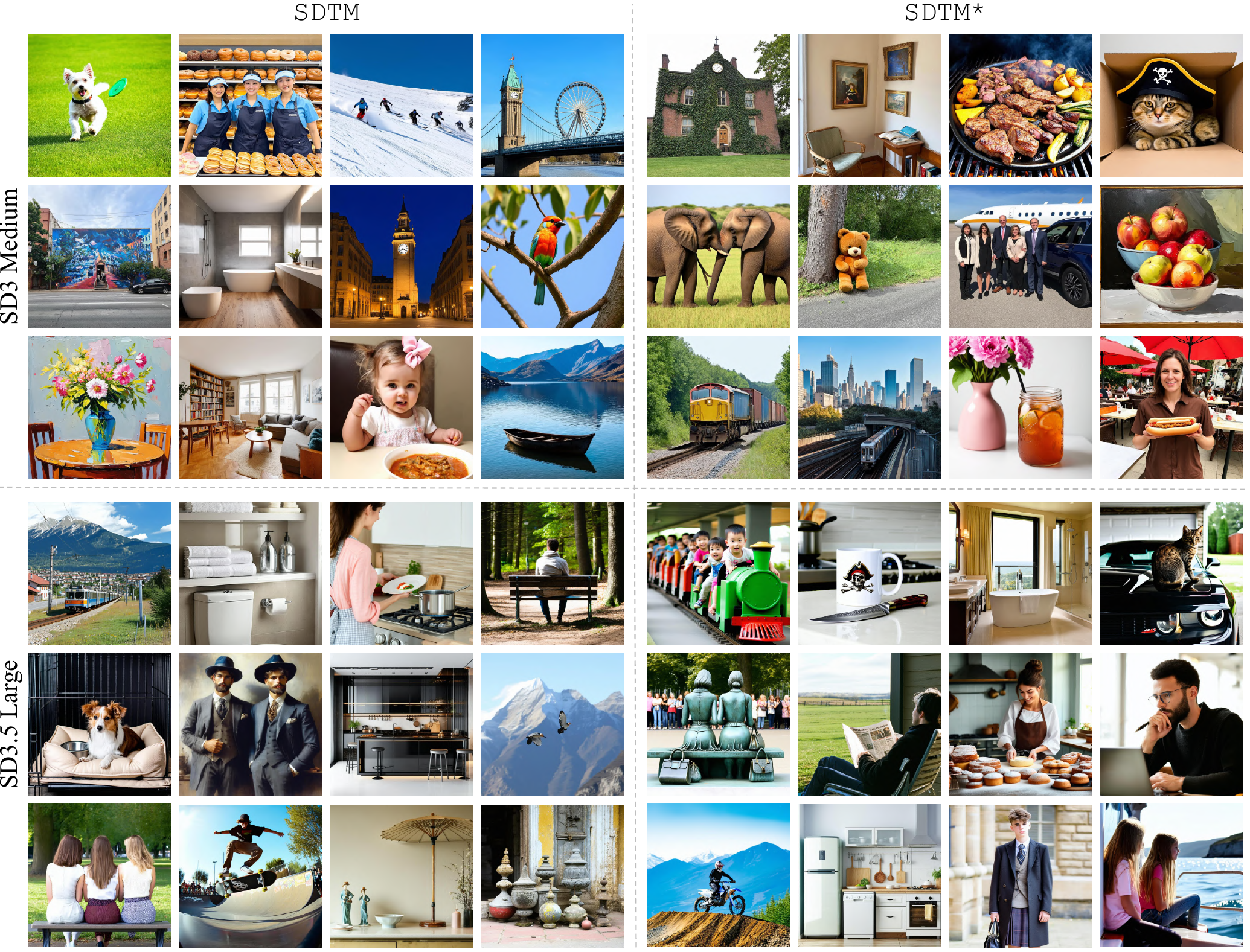}

	\caption{Uncurated images generated using SD3 Medium and SD3.5 Large configurations under the \texttt{SDTM} and \texttt{SDTM}* frameworks. } 
    \vspace{-12 pt}
	\label{fig:all}
\end{figure*}

\subsection{More Visualization of Samples}
In Fig.~\ref{fig:comparison_supp}, we compare our methods with fine-tuning-free techniques including ToMeSD and AT-EDM based on the SD3.5 Large. Fig.~\ref{fig:all} displays uncurated images generated by our methods across SD3 Medium and SD3.5 Large. Our methods outperform other SoTA techniques and maintain robust generative capabilities across various scenes.
\label{app_sec:qualitative}

\end{document}